# DeepVO: Towards End-to-End Visual Odometry with Deep Recurrent Convolutional Neural Networks

Sen Wang, Ronald Clark, Hongkai Wen and Niki Trigoni

***Abstract*— This paper studies monocular visual odometry (VO) problem. Most of existing VO algorithms are developed under a standard pipeline including feature extraction, feature matching, motion estimation, local optimisation, etc. Although some of them have demonstrated superior performance, they usually need to be carefully designed and specifically fine-tuned to work well in different environments. Some prior knowledge is also required to recover an absolute scale for monocular VO. This paper presents a novel end-to-end framework for monocular VO by using deep Recurrent Convolutional Neural Networks (RCNNs) [1]. Since it is trained and deployed in an end-to-end manner, it infers poses directly from a sequence of raw RGB images (videos) without adopting any module in the conventional VO pipeline. Based on the RCNNs, it not only automatically learns effective feature representation for the VO problem through Convolutional Neural Networks, but also implicitly models sequential dynamics and relations using deep Recurrent Neural Networks. Extensive experiments on the KITTI VO dataset show competitive performance to state-of-the-art methods, verifying that the end-to-end Deep Learning technique can be a viable complement to the traditional VO systems.**

## I. Introduction

Visual odometry (VO), as one of the most essential techniques for pose estimation and robot localisation, has attracted significant interest in both the computer vision and robotics communities over the past few decades [1]. It has been widely applied to various robots as a complement to GPS, Inertial Navigation System (INS), wheel odometry, etc.

In the last thirty years, enormous work has been done to develop an accurate and robust monocular VO system. As shown in Fig. 1, a classic pipeline [1], [2], which typically consists of camera calibration, feature detection, feature matching (or tracking), outlier rejection (e.g., RANSAC), motion estimation, scale estimation and local optimisation (Bundle Adjustment), has been developed and broadly recognised as a golden rule to follow. Although some state-of-the-art algorithms based on this pipeline have shown excellent performance in terms of accuracy and robustness, they are usually hard-coded with significant engineering effort and each module in the pipeline needs to be carefully designed and fine-tuned to ensure the performance. Moreover, the monocular VO has to estimate an absolute scale by using some extra information (e,g., height of the camera) or prior knowledge, making it prone to big drift and more challenging than the stereo VO.

The authors are with Department of Computer Science, University of Oxford, Oxford OX1 3QD, United Kingdom {firstname.lastname}@cs.ox.ac.uk

[1]Project Website: http://senwang.gitlab.io/DeepVO/

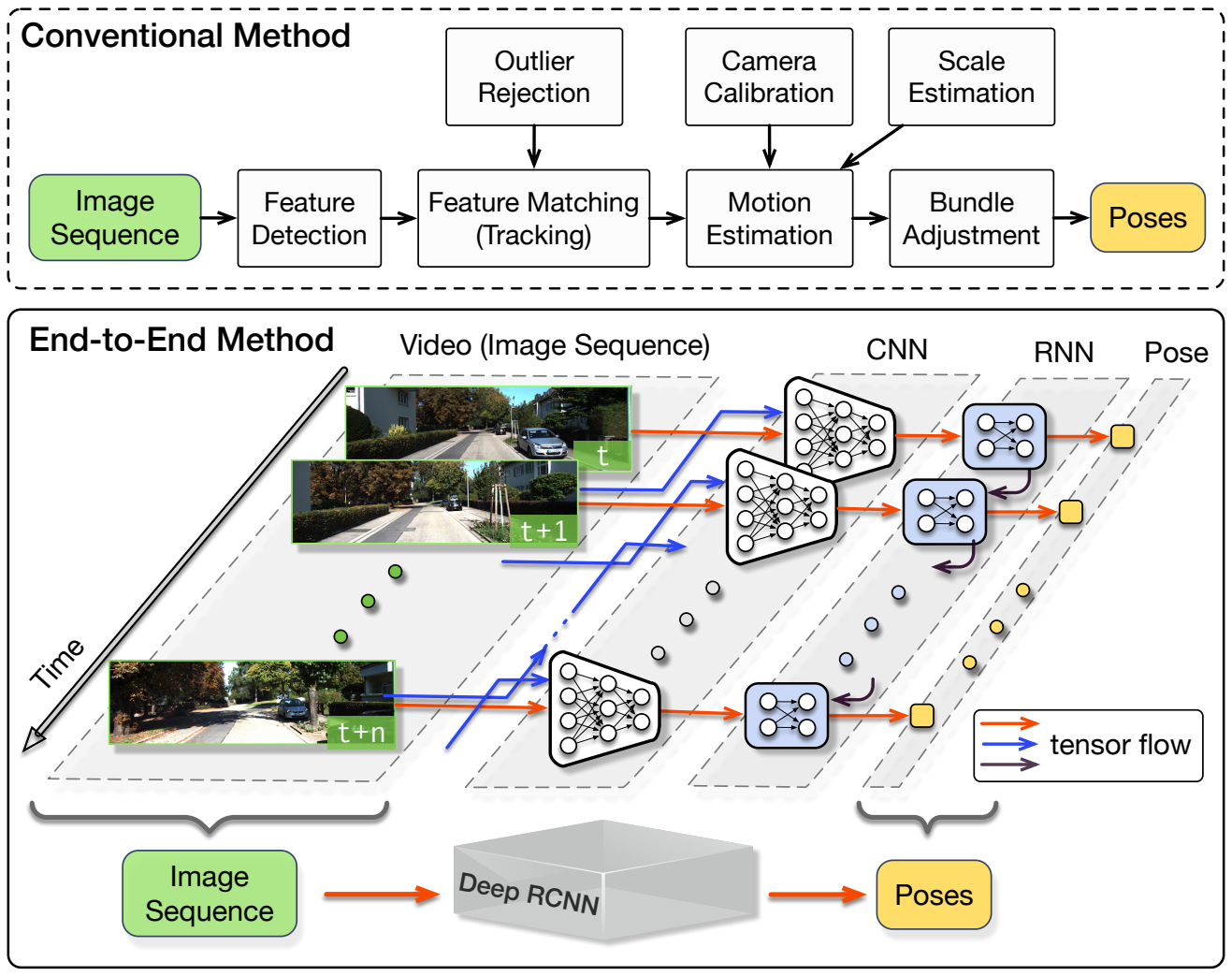

Fig. 1. Architectures of the conventional feature based monocular VO and the proposed end-to-end method. In the proposed method, RCNN takes a sequence of RGB images (video) as input and learns features by CNN for RNN based sequential modelling to estimate poses. Camera image credit: KITTI dataset [3].

Deep Learning (DL) has recently been dominating many computer vision tasks with promising results. Unfortunately, for the VO problem this has not arrived yet. In fact, there is very limited work on VO, even related to 3D geometry problems. We presume that this is because most of the existing DL architectures and pre-trained models are essentially designed to tackle recognition and classification problems, which drives deep Convolutional Neural Networks (CNNs) to extract high-level appearance information from images. Learning the appearance representation confines the VO to function only in trained environments and seriously hinders the popularisation of the VO to new scenarios. This is why the VO algorithms heavily rely on geometric features rather than appearance ones. Meanwhile, a VO algorithm ideally should model motion dynamics by examining the changes and connections on a sequence of images rather than processing a single image. This means we need sequential learning, which the CNNs are inadequate to.

In this paper, we propose a novel DL based monocular VO algorithm by leveraging deep Recurrent Convolutional Neural Networks (RCNNs) [4]. Since it is achieved in an end-to-end manner, it does not need any module in the classic VO pipeline (even camera calibration). The main contribution is threefold: 1) We demonstrate that the monocular VO problem can be addressed in an end-to-end fashion based on DL, i.e., directly estimating poses from raw RGB images. Neither prior knowledge nor parameter is needed to recover the absolute scale. To the best of our knowledge, this is the first end-to-end approach on the monocular VO through Deep Neural Networks (DNNs). 2) We propose a RCNN architecture enabling the DL based VO algorithm to be generalised to totally new environments by using the geometric feature representation learnt by the CNN. 3) Sequential dependence and complex motion dynamics of an image sequence, which are of importance to the VO but cannot be explicitly or easily modelled by human, are implicitly encapsulated and automatically learnt by deep Recurrent Neural Networks (RNNs).

The rest of this paper is organised as follows. Section II reviews related work. The end-to-end monocular VO algorithm is described in Section III, followed by experimental results in Section IV. Conclusion is drawn in Section V.

## II. Related Work

Early work on the monocular VO is reviewed in this section, discussing various algorithms and their differences from others. There are mainly two types of algorithms in terms of the technique and framework adopted: geometry based and learning based methods.

### A. Methods based on Geometry

Theoretically based on geometric theory, geometry based methods, which dominate the area of VO, rely on geometric constraints extracted from imagery to estimate motion. Since they are derived from elegant and established principles and have been extensively investigated, most of state-of-the-art VO algorithms fall into this family. They can be further divided into sparse feature based methods and direct methods.

*1) Sparse Feature based Methods:* Sparse feature based methods, whose typical pipeline is shown in Fig. 1, employ multi-view geometry [5] to determine motion after extracting and matching (or tracking) salient feature points from a sequence of images, such as the algorithm in [6] and LIBVISO2 [7]. However, due to the presence of outliers, noises, etc., all VO algorithms suffer from drifts over time. To mitigate this problem, visual Simultaneous Localisation and Mapping (SLAM) or Structure from Motion (SfM) can be adopted to maintain a feature map for drift correction along with pose estimation [8]. Examples include keyframe base PTAM [9] and ORB-SLAM [10].

*2) Direct Methods:* Feature extraction and matching of sparse feature based methods are computationally expensive. More importantly, they only use salient features without benefiting from rich information contained in the whole image. Direct methods, in contrast, are capable of exploiting all the pixels in consecutive images for pose estimation under the assumption of photometric consistency, e.g., DTAM in [11]. Recently, semi-direct approaches which realise superior performance are developed for the monocular VO [12]–[14]. Since the direct methods tend to be more accurate in principle than feature based ones and can work better in texture-less environments, they are increasingly gaining more favour.

### B. Methods based on Learning

As data-driven approaches, learning based methods are to learn motion model and infer VO from sensor readings by Machine Learning techniques without explicitly applying geometric theory. Optical flow is used to train K Nearest Neighbour (KNN), Gaussian Process (GP) and Support Vector Machines (SVM) regression algorithms for the monocular VO in [15], [16] and [17], respectively. Since the learning based methods are recently emerging, there is limited amount of work and no one has directly dealt with raw RGB images yet.

It has been widely recognised that traditional Machine Learning techniques are inefficient when encountering big or highly non-linear, high-dimensional data, e.g., RGB images. DL which automatically learns suitable feature representation from large-scale dataset provides an alternative solution to the VO problem.

*1) Deep Learning based Methods:* DL has achieved promising results on some localisation related applications. The features of CNNs, for instance, have been utilised for appearance based place recognition [18]. Unfortunately, there is little work on VO or pose estimation. To our knowledge, [19] firstly realises DL based VO through synchrony detection between image sequences and features. After estimating depth from stereo images, the CNN predicts the discretised changes of direction and velocity by the softmax function. Although this work provides a feasible scheme for DL based stereo VO, it inherently formulates the VO as a classification problem rather than pose regression. Camera relocalisation using a single image is solved in [20] by fine-tuning images of a specific scene with CNNs. It suggests to label these images by SfM, which is time-consuming and labour-intensive for large-scale scenarios. Because a trained CNN model serves as an appearance "map" of the scene, it needs to be re-trained or at least fine-tuned for a new environment. This seriously hampers the technique for widespread usage, which is also one of the biggest difficulties when applying DL for VO. To overcome this problem, the CNNs are provided with dense optical flow instead of RGB images for motion estimation in [21]. Three different architectures of CNNs are developed to learn appropriate features for VO, achieving robust VO even with blurred and under-exposured images. However, the proposed CNNs require pre-processed dense optical flow as input, which cannot benefit from the end-to-end learning and may be inappropriate to real-time applications.

Because the CNNs are incapable of modelling sequential information, none of the previous work considers image sequences or videos for sequential learning. In this work, we tackle this by leveraging the RNNs.

## III. End-to-End Visual Odometry through RCNN

In this section, the deep RCNN framework realising the monocular VO in an end-to-end fashion is described in detail.

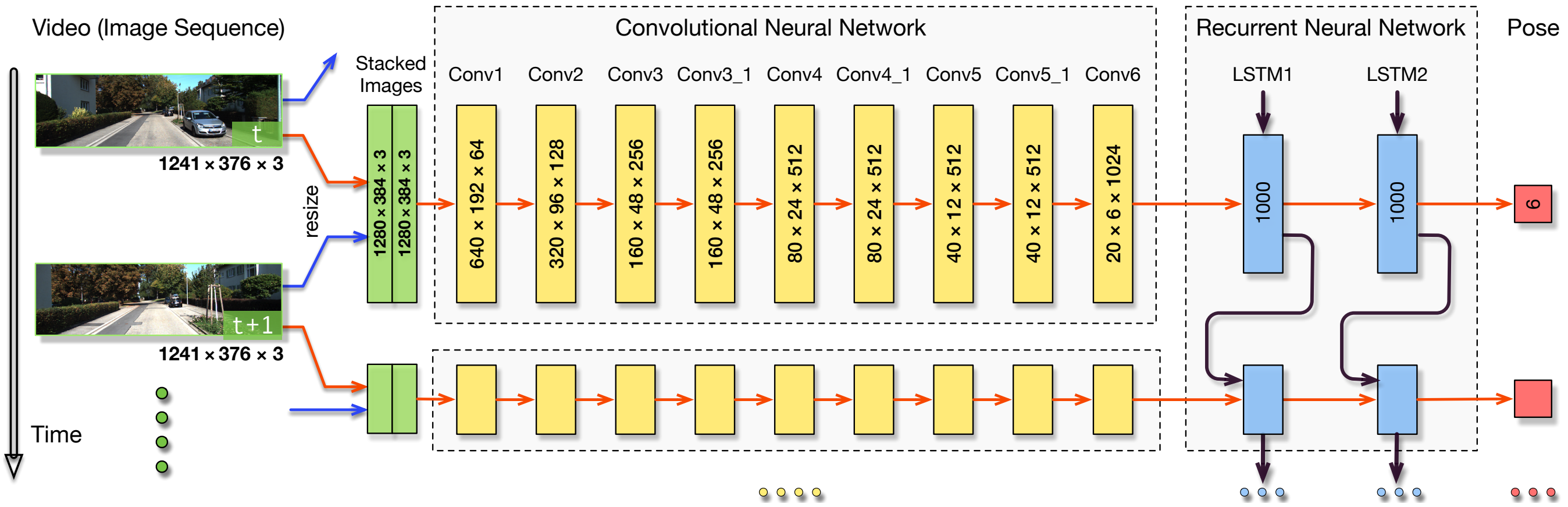

Fig. 2. Architecture of the proposed RCNN based monocular VO system. The dimensions of the tensors shown here are given as an example based on the image size of the KITTI dataset. The CNN ones should vary according to the size of the input image. Camera image credit: KITTI dataset.

It is mainly composed of CNN based feature extraction and RNN based sequential modelling.

### A. *Architecture of the Proposed RCNN*

There have been some popular and powerful DNN architectures, such as VGGNet [22] and GoogLeNet [23], developed for computer vision tasks, producing remarkable performance. Most of them are designed with tackling recognition, classification and detection problems in mind, which means that they are trained to learn knowledge from appearance and image context. However, as discussed before, VO which is rooted in geometry should not be closely coupled with appearance. Therefore, it is impractical to simply adopt the current popular DNN architectures for the VO problem. A framework which can learn geometric feature representations is of importance to address the VO and other geometric problems. Meanwhile, it is essential to derive connections among consecutive image frames, e.g., motion models, since the VO systems evolve over time and operate on image sequences acquired during movement. Therefore, the proposed RCNN takes these two requirements into consideration.

The architecture of the proposed end-to-end VO system is shown in Fig. 2. It takes a video clip or a monocular image sequence as input. At each time step, the RGB image frame is pre-processed by subtracting the mean RGB values of the training set and, optionally, resizing to a new size in the multiple of 64. Two consecutive images are stacked together to form a tensor for the deep RCNN to learn how to extract motion information and estimate poses. Specifically, the image tensor is fed into the CNN to produce an effective feature for the monocular VO, which is then passed through a RNN for sequential learning. Each image pair yields a pose estimate at each time step through the network. The VO system develops over time and estimates new poses as images are captured.

The advantage of the RCNN based architecture is to allow simultaneous feature extraction and sequential modelling of VO through the combination of CNN and RNN. More details are given in the subsequent sections.

TABLE I
CONFIGURATION OF THE CNN

| Layer | Receptive Field Size | Padding | Stride | Number of Channels |
|---|---|---|---|---|
| Conv1 | $7 \times 7$ | 3 | 2 | 64 |
| Conv2 | $5 \times 5$ | 2 | 2 | 128 |
| Conv3 | $5 \times 5$ | 2 | 2 | 256 |
| Conv3_1 | $3 \times 3$ | 1 | 1 | 256 |
| Conv4 | $3 \times 3$ | 1 | 2 | 512 |
| Conv4_1 | $3 \times 3$ | 1 | 1 | 512 |
| Conv5 | $3 \times 3$ | 1 | 2 | 512 |
| Conv5_1 | $3 \times 3$ | 1 | 1 | 512 |
| Conv6 | $3 \times 3$ | 1 | 2 | 1024 |

### B. *CNN based Feature Extraction*

In order to automatically learn effective features that are suitable for the VO problem, a CNN is developed to perform feature extraction on the concatenation of two consecutive monocular RGB images. The feature representation is ideally geometric instead of being associated with appearance or visual context because the VO systems need to be generalised and deployed in unknown environments. The structure of the CNN is inspired by the network for optical flow estimation in [24].

The configuration of the CNN is outlined in TABLE I and an example of its tensors on KITTI dataset is given in Fig. 2. It has 9 convolutional layers and each layer is followed by a rectified linear unit (ReLU) activation except `Conv6`, i.e., 17 layers in total. The sizes of the receptive fields in the network gradually reduce from $7 \times 7$ to $5 \times 5$ and then $3 \times 3$ to capture small interesting features. Zero-paddings are introduced to either adapt to the configurations of the receptive fields or preserve the spatial dimension of the tensor after convolution. The number of the channels, i.e., the number of filters for feature detection, increases to learn various features.

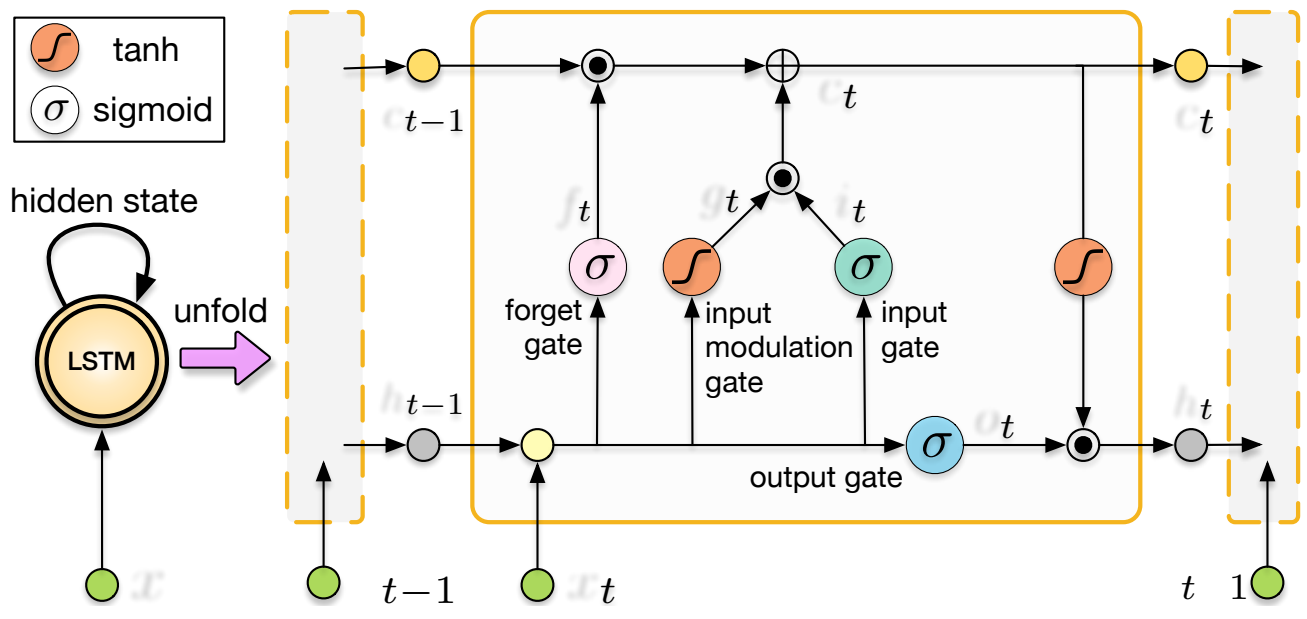

Fig. 3. Folded and unfolded LSTMs and internal structure of its unit. $\odot$ and $\oplus$ denote element-wise product and addition of two vectors, respectively.

The CNN takes raw RGB images instead of pre-processed counterparts, such as optical flow or depth images, as input because the network is trained to learn an efficient feature representation with reduced dimensionality for the VO. This learnt feature representation not only compresses the original high-dimensional RGB image into a compact description, but also boosts the successive sequential training procedure. Hence, the last convolutional feature `Conv6` is passed to the RNN for sequential modelling.

### C. RNN based Sequential Modelling

Following the CNN, a deep RNN is designed to conduct sequential learning, i.e., to model dynamics and relations among a sequence of CNN features. Note that this modelling is performed implicitly by the RNN to automatically discover appropriate sequential knowledge. Therefore, it may come out more than the models we use to describe physical movement and geometry.

Since the RNN is capable of modelling dependencies in a sequence, it is well suited to the VO problem which involves temporal model (motion model) and sequential data (image sequence). For instance, estimating pose of current image frame can benefit from information encapsulated in previous frames. In fact, this insight has already existed in the conventional VO systems. For example, multi-view geometry is able to avoid some issues in two-view geometry [5]. However, RNN is not suitable to directly learn sequential representation from high-dimensional raw data, such as images. Therefore, the proposed system adopts the appealing RCNN architecture with the CNN features as the input of the RNN.

RNN is different from CNN in that it maintains memory of its hidden states over time and has feedback loops among them, which enables its current hidden state to be a function of the previous ones, as the RNN part shown in Fig. 2. Hence, the RNN can find out the connections among the input and the previous states in the sequence. Given a convolutional feature $\mathbf{x}_k$ at time $k$, a RNN updates at time step $k$ by

$$\begin{aligned} \mathbf{h}_k &= \mathcal{H}(\mathbf{W}_{xh}\mathbf{x}_k + \mathbf{W}_{hh}\mathbf{h}_{k-1} + \mathbf{b}_h) \\ \mathbf{y}_k &= \mathbf{W}_{hy}\mathbf{h}_k + \mathbf{b}_y \end{aligned} \tag{1}$$

where $\mathbf{h}_k$ and $\mathbf{y}_k$ are the hidden state and output at time $k$ respectively, $\mathbf{W}$ terms denote corresponding weight matrices, $b$ terms denote bias vectors, and $\mathcal{H}$ is an element-wise non-linear activation function, such as sigmoid or hyperbolic tangent. Although in theory the standard RNN can learn sequences with arbitrary lengths, it is limited to short ones in practice due to the known vanishing gradient problem [25].

In order to be able to find and exploit correlations among images taken in long trajectories, Long Short-Term Memory (LSTM) which is capable of learning long-term dependencies by introducing memory gates and units [26] is employed as our RNN. It explicitly determines which previous hidden states to be discarded or retained for updating the current state, being expected to learn the motion during pose estimation. The folded LSTM and its unfolded version over time are shown in Fig. 3 along with the internal structure of a LSTM unit. It can be seen that after unfolding the LSTM, each LSTM unit is associated with a time step. Given the input $\mathbf{x}_k$ at time $k$ and the hidden state $\mathbf{h}_{k-1}$ and the memory cell $\mathbf{c}_{k-1}$ of the previous LSTM unit, the LSTM updates at time step $k$ according to

$$\begin{aligned} \mathbf{i}_k &= \sigma(\mathbf{W}_{xi}\mathbf{x}_k + \mathbf{W}_{hi}\mathbf{h}_{k-1} + \mathbf{b}_i) \\ \mathbf{f}_k &= \sigma(\mathbf{W}_{xf}\mathbf{x}_k + \mathbf{W}_{hf}\mathbf{h}_{k-1} + \mathbf{b}_f) \\ \mathbf{g}_k &= \tanh(\mathbf{W}_{xg}\mathbf{x}_k + \mathbf{W}_{hg}\mathbf{h}_{k-1} + \mathbf{b}_g) \\ \mathbf{c}_k &= \mathbf{f}_k \odot \mathbf{c}_{k-1} + \mathbf{i}_k \odot \mathbf{g}_k \\ \mathbf{o}_k &= \sigma(\mathbf{W}_{xo}\mathbf{x}_k + \mathbf{W}_{ho}\mathbf{h}_{k-1} + \mathbf{b}_o) \\ \mathbf{h}_k &= \mathbf{o}_k \odot \tanh(\mathbf{c}_k) \end{aligned} \tag{2}$$

where $\odot$ is element-wise product of two vectors, $\sigma$ is sigmoid non-linearity, $\tanh$ is hyperbolic tangent non-linearity, $\mathbf{W}$ terms denote corresponding weight matrices, $b$ terms denote bias vectors, $\mathbf{i}_k$, $\mathbf{f}_k$, $\mathbf{g}_k$, $\mathbf{c}_k$ and $\mathbf{o}_k$ are input gate, forget gate, input modulation gate, memory cell and output gate at time $k$, respectively,

Although the LSTM can handle long-term dependencies and has deep temporal structure, it still needs depth on network layers to learn high level representation and model complex dynamics. The advantages of the deep RNN architecture have been proved in [27] for speech recognition using acoustic signal. Therefore, in our case the deep RNN is constructed by stacking two LSTM layers with the hidden states of a LSTM being the input of the other one, as illustrated in Fig. 2. In our network, each of the LSTM layers has 1000 hidden states.

The deep RNN outputs a pose estimate at each time step based on the visual features generated from the CNN. This progresses over time as the camera moves and images are captured.

### D. Cost Function and Optimisation

The proposed RCNN based VO system can be considered to compute the conditional probability of the poses $\mathbf{Y}_t = (\mathbf{y}_1, \ldots, \mathbf{y}_t)$ given a sequence of monocular RGB images $\mathbf{X}_t = (\mathbf{x}_1, \ldots, \mathbf{x}_t)$ up to time $t$ in the probabilistic perspective:

$$p(\mathbf{Y}_t|\mathbf{X}_t) = p(\mathbf{y}_1, \ldots, \mathbf{y}_t|\mathbf{x}_1, \ldots, \mathbf{x}_t) \tag{3}$$

The modelling and probabilistic inference are performed in the deep RCNN. To find the optimal parameters $\boldsymbol{\theta}^*$ for the

VO, the DNN maximises (3):

$$\boldsymbol{\theta}^* = \underset{\boldsymbol{\theta}}{\operatorname{argmax}}\, p(\mathbf{Y}_t|\mathbf{X}_t;\boldsymbol{\theta}) \tag{4}$$

To learn the hyperparameters $\boldsymbol{\theta}$ of the DNNs, the Euclidean distance between the ground truth pose $(\mathbf{p}_k, \boldsymbol{\varphi}_k)$ at time $k$ and its estimated one $(\widehat{\mathbf{p}}_k, \widehat{\boldsymbol{\varphi}}_k)$ is minimised. The loss function is composed of Mean Square Error (MSE) of all positions $\mathbf{p}$ and orientations $\boldsymbol{\varphi}$:

$$\boldsymbol{\theta}^* = \underset{\boldsymbol{\theta}}{\operatorname{argmin}} \frac{1}{N}\sum_{i=1}^{N}\sum_{k=1}^{t} \|\widehat{\mathbf{p}}_k - \mathbf{p}_k\|_2^2 + \kappa\|\widehat{\boldsymbol{\varphi}}_k - \boldsymbol{\varphi}_k\|_2^2 \tag{5}$$

where $\|\cdot\|$ is 2-norm, $\kappa$ (100 in the experiments) is a scale factor to balance the weights of positions and orientations, and $N$ is the number of samples. The orientation $\boldsymbol{\varphi}$ is represented by Euler angles rather than quaternion since quaternion is subject to an extra unit constraint which hinders the optimisation problem of DL. We also find that in practice using quaternion degrades the orientation estimate to some extent.

## IV. Experimental Results

In this section, we evaluate the proposed end-to-end monocular VO approach on the well-known KITTI VO/SLAM benchmark [3]. Since most of existing monocular VO algorithms do not estimate an absolute scale, their localisation results have to be manually aligned with ground truth. Therefore, the open-source VO library LIBVISO2 [7] which uses a fixed camera height to recover the scale for the monocular VO is adopted for comparison. Its stereo version which can directly obtain the absolute poses is also employed.

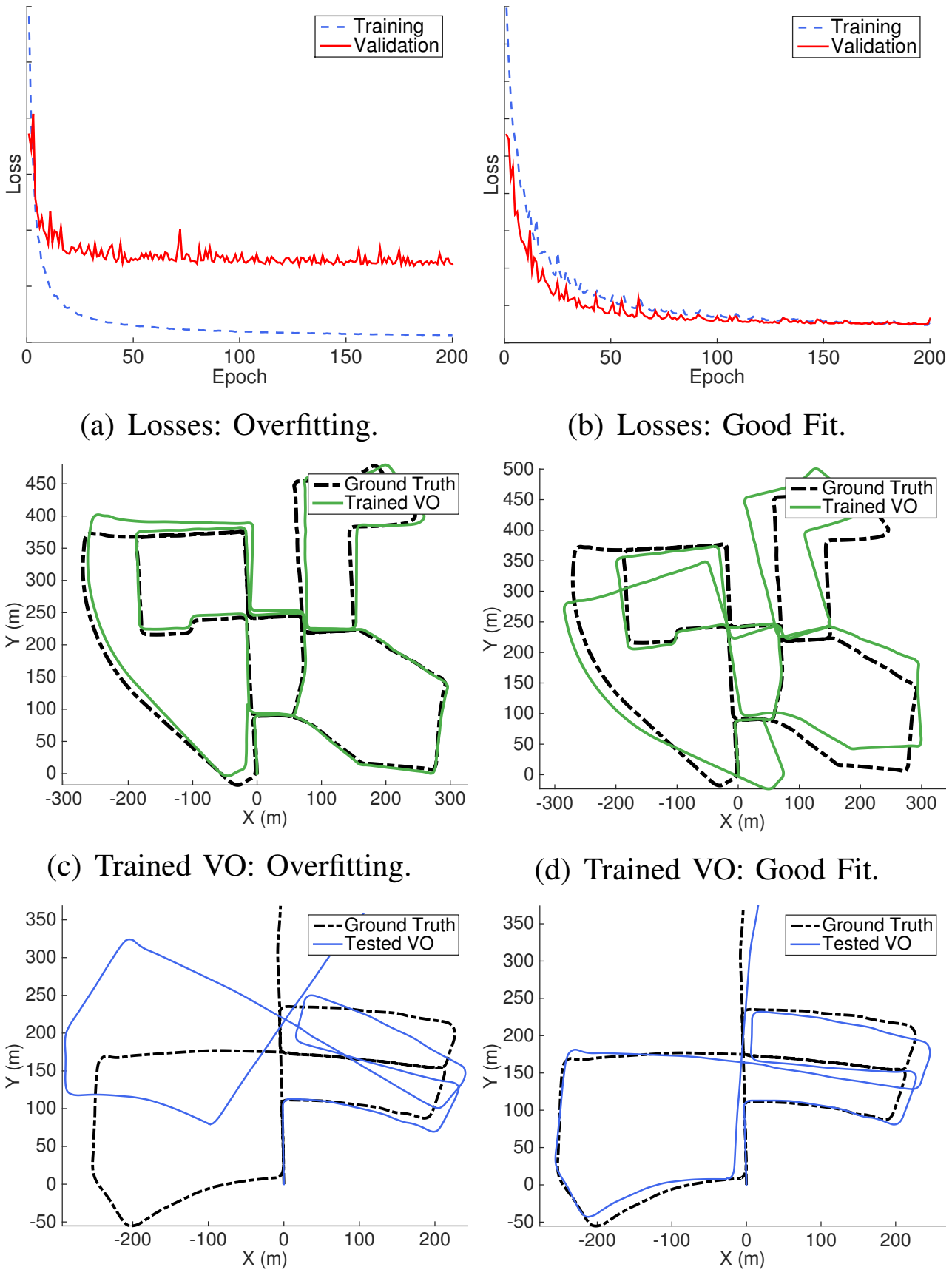

(a) Losses: Overfitting. (b) Losses: Good Fit.
(c) Trained VO: Overfitting. (d) Trained VO: Good Fit.
(e) Tested VO: Overfitting. (f) Tested VO: Good Fit.

Fig. 4. Training losses and VO results of two models. Figures in the left and right columns are about the over-fitted and well-fitted models, respectively. (a)-(b) Training and validation losses. (c)-(d) Estimated VO on training data (Sequence 00). (e)-(f) Estimated VO on testing data (Sequence 05).

### A. Training and Testing

*1) Dataset:* The KITTI VO/SLAM benchmark [3] has 22 sequences of images, of which 11 ones (Sequence 00-10) are associated with ground truth. The other 10 sequences (Sequence 11-21) are only provided with raw sensor data. Since this dataset was recorded at a relatively low frame rate (10 fps) by driving in urban areas with many dynamic objects and the speed of the driving was up to 90 km/h, it is very challenging for the monocular VO algorithms.

*2) Training and Testing:* Two separate experiments are conducted to evaluate the proposed method. The first one is based on the Sequence 00-10 to quantitatively analyse its performance by ground truth since the ground truth is only provided for these sequences. In order to have a separate dataset for testing, only the Sequence 00, 02, 08 and 09 which are relatively long are used for training. The trajectories are segmented to different lengths to generate much data for training, producing 7410 samples in total. The trained models are tested on the Sequence 03, 04, 05, 06, 07 and 10 for evaluation.

Since the ability to generalise well to real data is essential for DL based approaches, the next experiment aims to analyse how the proposed method and the trained VO models behave in totally new environments. For the VO problem, this is further required as aforementioned. Therefore, models trained on all the Sequence 00-10 are tested on the Sequence 11-21 which do not have ground truth available for training.

The network is implemented based on the famous DL framework Theano and trained by using a NVIDIA Tesla K40 GPU. The Adagrad optimiser is employed to train the network for up to 200 epochs with learning rate 0.001. Dropout and early stopping techniques are introduced to prevent the models from overfitting. In order to reduce both the training time and data required to converge, the CNN is based on a pre-trained FlowNet model [24].

*3) How overfitting affects the VO:* It is well known that overfitting is an undesirable behaviour for Machine Learning based methods. However, its meaning and influence are unclear in the context of the VO problem. Concrete discussions on this, which can guide a better training on the VO system, are still missing. Some insights on our training procedure and results are described here. In Fig. 4, the losses and VO results of two models are given. The big gap between the training and validation losses in Fig. 4(a) indicates serious overfitting compared to the proper losses in Fig. 4(b). Reflecting on the estimated VO of the training data, the results of the over-fitted model are much more accurate that those of the well-fitted model, as shown in Fig. 4(c) and Fig. 4(d). However, when applying the trained models on the testing data, the well-fitted model yields much better results, see Fig. 4(e) and Fig. 4(f). This is also very likely to happen when the model is deployed in practice working on real data. Therefore, overfitting should be carefully examined when training a model for the VO. Based on this example, it is can be seen that for the DL based VO problem overfitting has very intuitive outcomes and can seriously degrade the odometry estimation. A well-fitted model is key to ensuring good generalisation and reliable pose estimation of the trained VO models to untrained environments. In our work, we observed found that the orientation is more prone to overfitting than position. This could be because the orientation changes are usually smaller. In terms of underfitting, we assume this is rare because the capacity of the DNN is typically large and the size of training data tends to be limited.

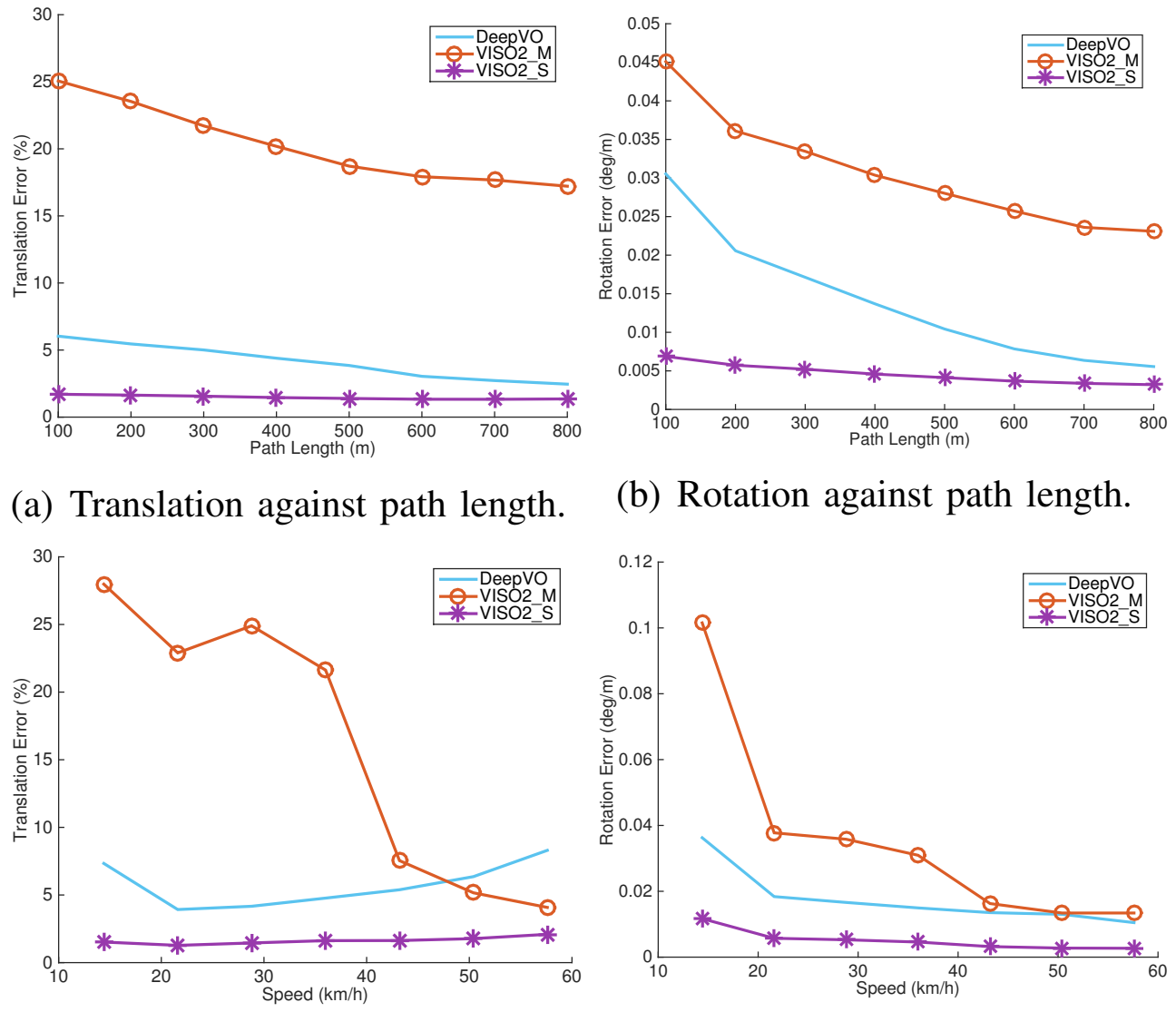

(a) Translation against path length. (b) Rotation against path length.

(c) Translation against speed. (d) Rotation against speed.

Fig. 5. Average errors on translation and rotation against different path lengths and speeds. The DeepVO model used is trained on Sequence 00, 02, 08 and 09.

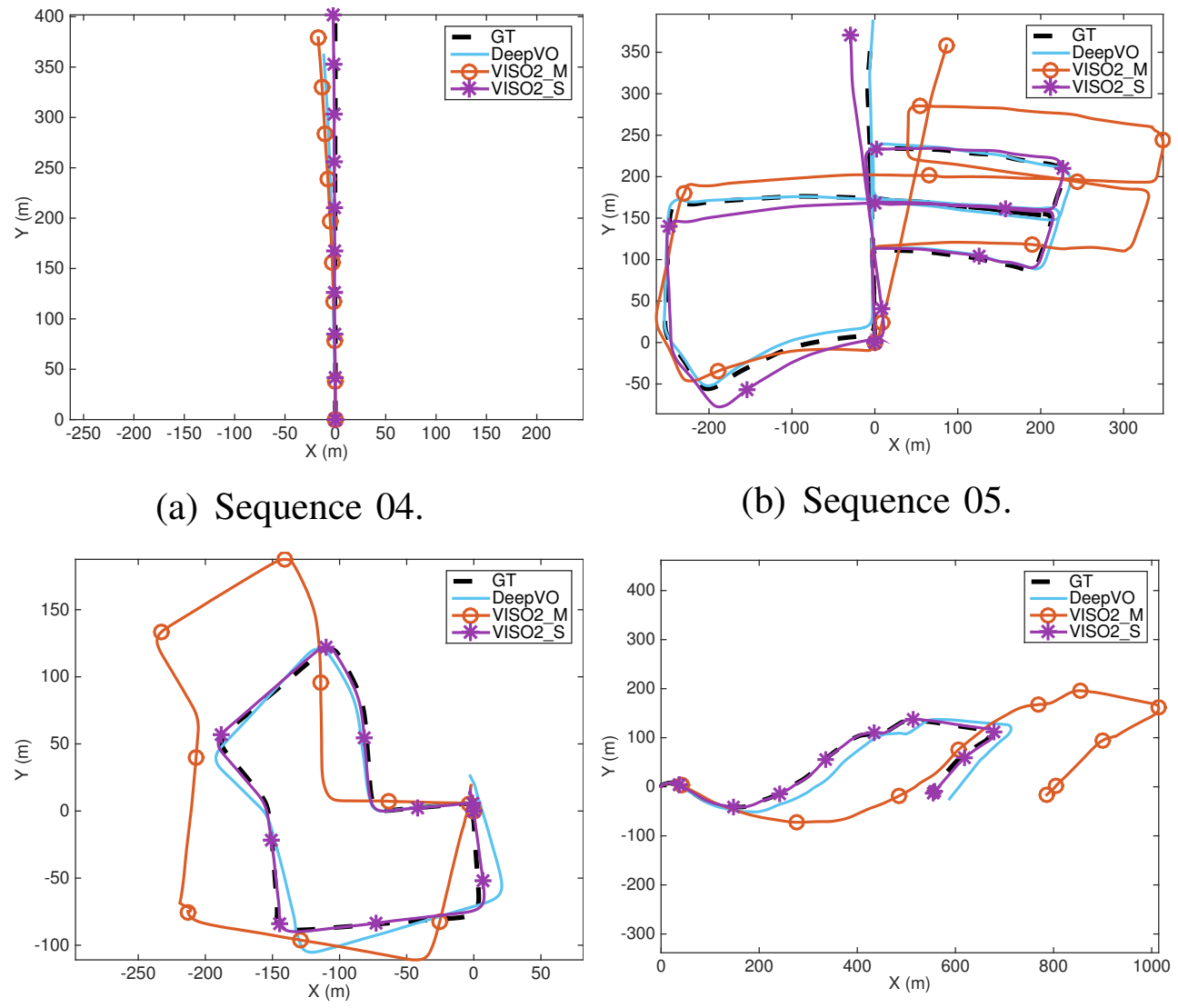

(a) Sequence 04. (b) Sequence 05.

(c) Sequence 07. (d) Sequence 10.

Fig. 6. Trajectories of VO testing results on Sequence 04, 05, 07 and 10. The DeepVO model used is trained on Sequence 00, 02, 08 and 09.

## B. VO Results

The performance of the trained VO models is analysed according to the KITTI VO/SLAM evaluation metrics, i.e., averaged Root Mean Square Errors (RMSEs) of the translational and rotational errors for all subsequences of lengths ranging from 100 to 800 meters and different speeds (the range of speeds varies in different sequences).

The first DL based model is trained on Sequence 00, 02, 08 and 09 and then tested on Sequence 03, 04, 05, 06, 07 and 10. The average RMSEs of the estimated VO on the test sequences are given in Fig. 5 with the translation and rotation against different path lengths and speeds. Although the result of the DeepVO is worst than that of the stereo VISO2 (VISO2_S), it is consistently better than the monocular VISO2 (VISO2_M) except for the translational errors of the DL model at high speeds, which are slightly higher than the monocular VISO2. We presume that this is because the maximum velocity of the Sequence 00, 02, 08 and 09 is below 60 km/h and there is very limited number of training samples whose speeds are bigger than 50 km/h. Without being trained with enough data covering the high-speed situation, the network tries to regress the VO but probably suffers from high drifts. It is interesting that the rotational errors become smaller on high velocities, which is opposite to the translation. This may be due to the fact that the KITTI dataset was recorded during car driving, which tends to go straight on high speeds yet rotate when slowing down. Moving forward, as a dynamics without significant changes on rotation, can be easily learnt to model by the RNN in terms of orientation, but the velocity varies fast. As the length of the trajectory increases, the errors of both the translation and rotation of the DeepVO significantly decrease, approaching to the stereo VISO2 as shown in Fig. 5(a) and Fig. 5(b).

The estimated VO trajectories corresponding to the previous testing are given in Fig. 6. It can be seen that the DeepVO produces relatively accurate and consistent trajectories against to the ground truth, demonstrating that the scale can be better estimated than using prior information, such as camera height. Note that no scale estimation or post alignment to ground truth is performed for the DeepVO to obtain the absolute poses. The scale is completely maintained by the network itself and implicitly learnt during the end-to-end training. Since recovering accurate and robust scale is surprisingly difficult for the monocular VO, this suggests an appealing advantage of the DL based VO method. The detailed performance of the algorithms on the testing sequences is summarised in TABLE II. It indicates that the DeepVO achieves more robust results than the monocular VISO2.

Although the generalisation of the DeepVO model has

TABLE II

RESULTS ON TESTING SEQUENCES.

| Seq. | DeepVO | | VISO2_M | | VISO2_S | |
|---|---|---|---|---|---|---|
| | $t_{\text{rel}}(\%)$ | $r_{\text{rel}}(^\circ)$ | $t_{\text{rel}}(\%)$ | $r_{\text{rel}}(^\circ)$ | $t_{\text{rel}}(\%)$ | $r_{\text{rel}}(^\circ)$ |
| 03 | 8.49 | 6.89 | 8.47 | 8.82 | 3.21 | 3.25 |
| 04 | 7.19 | 6.97 | 4.69 | 4.49 | 2.12 | 2.12 |
| 05 | 2.62 | 3.61 | 19.22 | 17.58 | 1.53 | 1.60 |
| 06 | 5.42 | 5.82 | 7.30 | 6.14 | 1.48 | 1.58 |
| 07 | 3.91 | 4.60 | 23.61 | 29.11 | 1.85 | 1.91 |
| 10 | 8.11 | 8.83 | 41.56 | 32.99 | 1.17 | 1.30 |
| mean | 5.96 | 6.12 | 17.48 | 16.52 | 1.89 | 1.96 |

- $t_{\text{rel}}$: average translational RMSE drift $(\%)$ on length of 100m-800m.
- $r_{\text{rel}}$: average rotational RMSE drift $(^\circ/100m)$ on length of 100m-800m.
- The DeepVO model used is trained on Sequence 00, 02, 08 and 09. Its performance is expected to improve when it is trained on more data.

been evaluated in the previous experiment, in order to further investigate how it performs in entirely new scenarios with different motion patterns and scenes, the network is tested on the testing dataset of KITTI VO benchmark. The DeepVO model is trained on all the 11 training sequences of the KITTI VO benchmark (i.e., Sequence 00-10), providing more data to avoid overfitting and to maximise the performance of the network. Due to the lack of ground truth for these testing sequences, no quantitative analysis can be performed on the VO results. For qualitative comparison, some predicted trajectories of the DeepVO, the monocular VISO2 and the stereo VISO2 are shown in Fig. 8. It can be seen that the results of the DeepVO are much better than these of the monocular VISO2 and roughly similar to the stereo VISO2's. It seems that this larger training dataset boosts the performance of the DeepVO. Taking the stereo properties of the stereo VISO2 into consideration, the DeepVO, as a monocular VO algorithm, achieves an appealing performance, showing that the trained model can generalise well in unknown scenarios. An exception could be the test on Sequence 12 in Fig. 8(b) which suffers from rather high localisation errors although the shape of the trajectory is close to the stereo VISO2's. There are several reasons. First, the training dataset does not have enough data on high speeds. Among all the 11 training dataset, only the Sequence 01 has velocities that are higher than 60 km/h. However, the speeds of the Sequence 12 span from 50km/h up to about 90km/h. Moreover, the images are captured at only 10 Hz, which makes the VO estimation more challenging during fast movement. The large open area around highway (lacking of features) and dynamic moving objects, shown in Fig. 7, can degrade the accuracy as well. These reasons also apply to Sequence 21. In order to mitigate these issues, the conventional geometry based methods could increase feature matching and introduce outlier rejection, such as RANSAC. However, for the DL based method, it is unclear how to embed these techniques yet. However, a feasible solution is to train the network with more data which not only reflects these situations but also is deliberatively augmented with noise, outliers, etc., allowing the network itself to figure out how to deal with these problems.

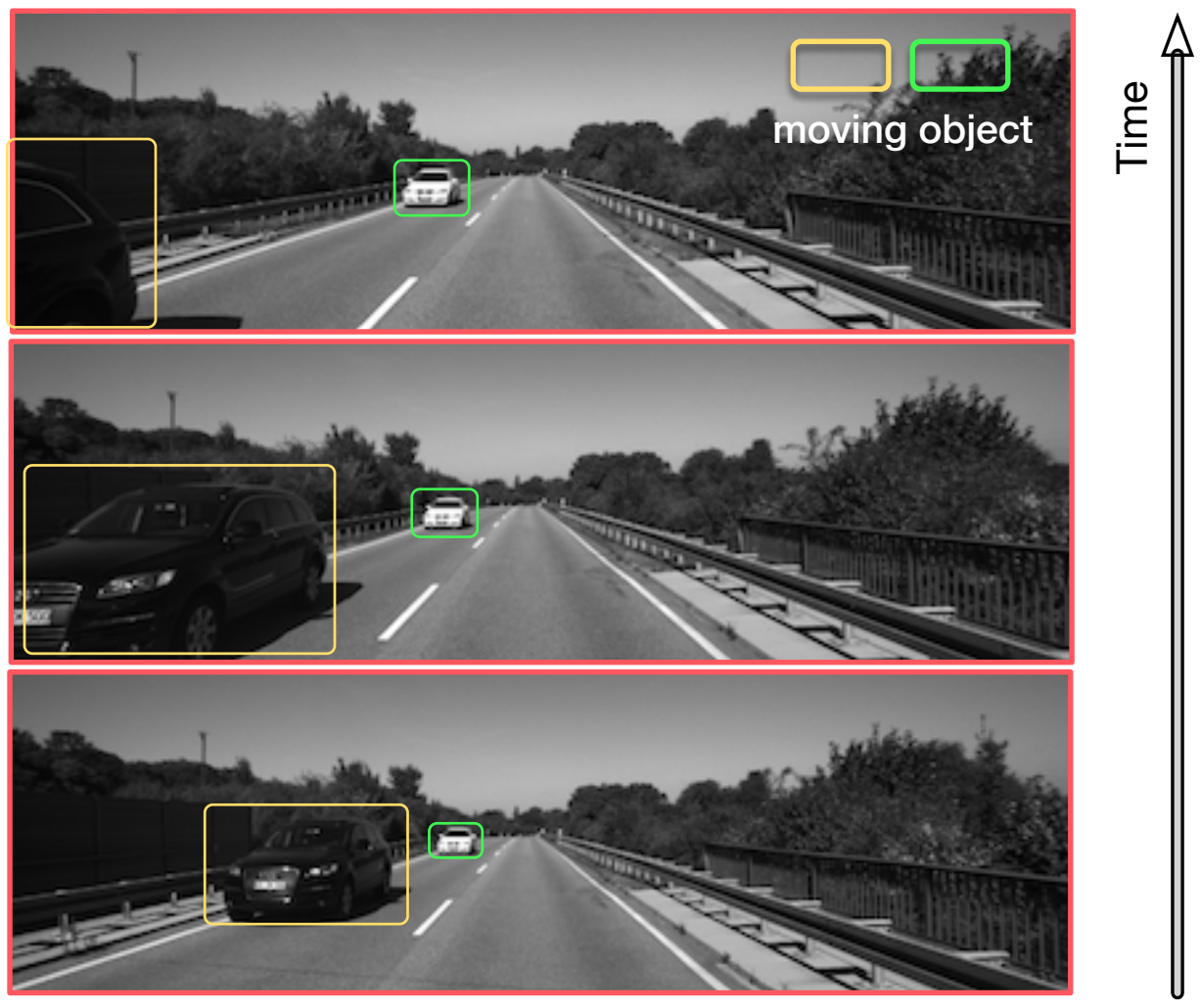

Fig. 7. Sample images from Sequence 12. There are some moving objects and large open areas in this sequence.

## V. CONCLUSIONS

This paper presents a novel end-to-end monocular VO algorithm based on Deep Learning. Leveraging the power of Deep RCNNs, this new paradigm is able to achieve simultaneous representation learning and sequential modelling of the the monocular VO by combining the CNNs with the RNNs. Since it does not depend on any module in the conventional VO algorithms (even camera calibration) for pose estimation and it is trained in an end-to-end manner, there is no need to carefully tune the parameters of the VO system. Based on the KITTI VO benchmark, it is verified that it can produce accurate VO results with precise scales and work well in completely new scenarios.

Although the proposed DL based VO method presents some results on this area, we stress that it is not expected as a replacement to the classic geometry based approach. On the contrary, it can be a viable complement, i.e., incorporating geometry with the representation, knowledge and models learnt by the DNNs to further improve the VO in terms of accuracy and, more importantly, robustness.

## ACKNOWLEDGMENT

This work is supported by EPSRC Programme Grant "Mobile Robotics: Enabling a Pervasive Technology of the Future" (EP/M019918/1). The authors would like to acknowledge the use of the University of Oxford Advanced Research Computing (ARC) facility in carrying out this work (http://dx.doi.org/10.5281/zenodo.22558).

## REFERENCES

[1] D. Scaramuzza and F. Fraundorfer, "Visual odometry: Tutorial," *IEEE Robotics & Automation Magazine*, vol. 18, no. 4, pp. 80–92, 2011.

[2] F. Fraundorfer and D. Scaramuzza, "Visual odometry: Part II: Matching, robustness, optimization, and applications," *IEEE Robotics & Automation Magazine*, vol. 19, no. 2, pp. 78–90, 2012.

[3] A. Geiger, P. Lenz, and R. Urtasun, "Are we ready for autonomous driving? the KITTI vision benchmark suite," in *Proceedings of IEEE Conference on Computer Vision and Pattern Recognition (CVPR)*, 2012.

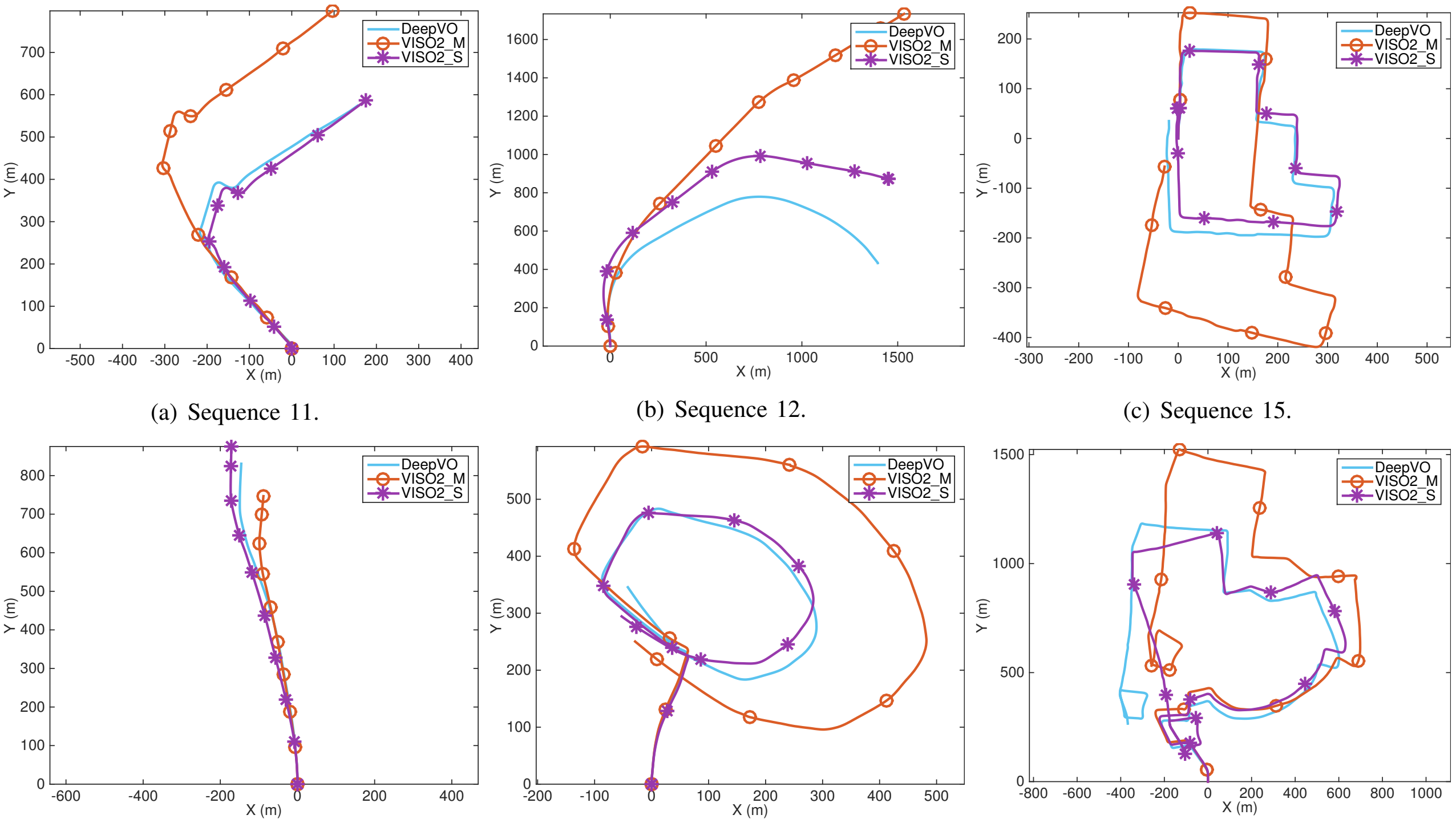

(a) Sequence 11. (b) Sequence 12. (c) Sequence 15.

(d) Sequence 17. (e) Sequence 18. (f) Sequence 19.

Fig. 8. Trajectories of VO results on the testing Sequence 11, 12, 15, 17, 18 and 19 of the KITTI VO benchmark (no ground truth is available for these testing sequences). The DeepVO model used is trained on the whole training dataset of the KITTI VO benchmark.

[4] J. Donahue, L. A. Hendricks, S. Guadarrama, M. Rohrbach, S. Venugopalan, K. Saenko, and T. Darrell, "Long-term recurrent convolutional networks for visual recognition and description," *IEEE Transactions on Pattern Analysis and Machine Intelligence*, to appear.

[5] R. Hartley and A. Zisserman, *Multiple view geometry in computer vision*. Cambridge university press, 2003.

[6] D. Nistér, O. Naroditsky, and J. Bergen, "Visual odometry," in *Proceedings of IEEE Conference on Computer Vision and Pattern Recognition (CVPR)*, vol. 1. IEEE, 2004, pp. I–652.

[7] A. Geiger, J. Ziegler, and C. Stiller, "Stereoscan: Dense 3D reconstruction in real-time," in *Intelligent Vehicles Symposium (IV)*, 2011.

[8] A. J. Davison, I. D. Reid, N. D. Molton, and O. Stasse, "MonoSLAM: Real-time single camera SLAM," *IEEE Transactions on Pattern Analysis and Machine Intelligence*, vol. 29, no. 6, pp. 1052–1067, 2007.

[9] G. Klein and D. Murray, "Parallel tracking and mapping for small AR workspaces," in *IEEE and ACM International Symposium on Mixed and Augmented Reality (ISMAR)*. IEEE, 2007, pp. 225–234.

[10] R. Mur-Artal, J. Montiel, and J. D. Tardós, "ORB-SLAM: a versatile and accurate monocular SLAM system," *IEEE Transactions on Robotics*, vol. 31, no. 5, pp. 1147–1163, 2015.

[11] R. A. Newcombe, S. J. Lovegrove, and A. J. Davison, "DTAM: Dense tracking and mapping in real-time," in *Proceedings of IEEE International Conference on Computer Vision (ICCV)*. IEEE, 2011, pp. 2320–2327.

[12] J. Engel, J. Sturm, and D. Cremers, "Semi-dense visual odometry for a monocular camera," in *Proceedings of IEEE International Conference on Computer Vision (ICCV)*, 2013, pp. 1449–1456.

[13] C. Forster, M. Pizzoli, and D. Scaramuzza, "SVO: Fast semi-direct monocular visual odometry," in *Proceedings of IEEE International Conference on Robotics and Automation (ICRA)*. IEEE, 2014, pp. 15–22.

[14] J. Engel, V. Koltun, and D. Cremers, "Direct sparse odometry," in *arXiv:1607.02565*, July 2016.

[15] R. Roberts, H. Nguyen, N. Krishnamurthi, and T. Balch, "Memory-based learning for visual odometry," in *Proceedings of IEEE International Conference on Robotics and Automation (ICRA)*. IEEE, 2008, pp. 47–52.

[16] V. Guizilini and F. Ramos, "Semi-parametric learning for visual odometry," *The International Journal of Robotics Research*, vol. 32, no. 5, pp. 526–546, 2013.

[17] T. A. Ciarfuglia, G. Costante, P. Valigi, and E. Ricci, "Evaluation of non-geometric methods for visual odometry," *Robotics and Autonomous Systems*, vol. 62, no. 12, pp. 1717–1730, 2014.

[18] N. Sünderhauf, S. Shirazi, A. Jacobson, F. Dayoub, E. Pepperell, B. Upcroft, and M. Milford, "Place recognition with convnet landmarks: Viewpoint-robust, condition-robust, training-free," in *Proceedings of Robotics: Science and Systems (RSS)*, 2015.

[19] K. Konda and R. Memisevic, "Learning visual odometry with a convolutional network," in *Proceedings of International Conference on Computer Vision Theory and Applications*, 2015.

[20] A. Kendall, M. Grimes, and R. Cipolla, "Convolutional networks for real-time 6-DoF camera relocalization," in *Proceedings of International Conference on Computer Vision (ICCV)*, 2015.

[21] G. Costante, M. Mancini, P. Valigi, and T. A. Ciarfuglia, "Exploring representation learning with CNNs for frame-to-frame ego-motion estimation," *IEEE Robotics and Automation Letters*, vol. 1, no. 1, pp. 18–25, 2016.

[22] K. Simonyan and A. Zisserman, "Very deep convolutional networks for large-scale image recognition," *arXiv preprint arXiv:1409.1556*, 2014.

[23] C. Szegedy, W. Liu, Y. Jia, P. Sermanet, S. Reed, D. Anguelov, D. Erhan, V. Vanhoucke, and A. Rabinovich, "Going deeper with convolutions," in *Proceedings of the IEEE Conference on Computer Vision and Pattern Recognition (CVPR)*, 2015, pp. 1–9.

[24] A. Dosovitskiy, P. Fischery, E. Ilg, C. Hazirbas, V. Golkov, P. van der Smagt, D. Cremers, T. Brox *et al.*, "Flownet: Learning optical flow with convolutional networks," in *Proceedings of IEEE International Conference on Computer Vision (ICCV)*. IEEE, 2015, pp. 2758–2766.

[25] I. Goodfellow, Y. Bengio, and A. Courville, "Deep learning," 2016, book in preparation for MIT Press.

[26] W. Zaremba and I. Sutskever, "Learning to execute," *arXiv preprint arXiv:1410.4615*, 2014.

[27] A. Graves and N. Jaitly, "Towards end-to-end speech recognition with recurrent neural networks." in *Proceedings of International Conference on Machine Learning (ICML)*, vol. 14, 2014, pp. 1764–1772.